%% file: main.tex
\documentclass[sigconf]{acmart}
\usepackage[nolist,nohyperlinks]{acronym} %Compatible
\usepackage{multirow} %Compatible
\usepackage{pifont} %Compatible
\usepackage{subcaption} %Compatible
\usepackage{mathtools} %Compatible

\AtBeginDocument{%
  \providecommand\BibTeX{{%
    \normalfont B\kern-0.5em{\scshape i\kern-0.25em b}\kern-0.8em\TeX}}}

\copyrightyear{2025}
\acmYear{2025}
\setcopyright{rightsretained}
\acmConference[GeoIndustry '25]{4th ACM SIGSPATIAL International Workshop on Spatial Big Data and AI for Industrial Applications}{November 3, 2025}{Minneapolis, MN, USA}
\acmBooktitle{4th ACM SIGSPATIAL International Workshop on Spatial Big Data and AI for Industrial Applications (GeoIndustry '25), November 3, 2025, Minneapolis, MN, USA}
\begin{document}

% ================ Title & Authors ================
\title{LocaGen: Low-Overhead Indoor Localization Through Spatial Augmentation}

\author{Abdelrahman Abdelmotlb}
\authornote{Equal contribution.}
\email{es-abdulrahmanm.mohammed2025@alexu.edu.eg}
\orcid{0009-0004-0416-7075}
\affiliation{%
   \institution{Alexandria University}
  \city{Alexandria}
  \country{Egypt}
}

\author{Abdallah Taman}
\authornotemark[1]
\email{es-abdullahm.abdullah2025@alexu.edu.eg}
\orcid{0009-0007-5239-7571}
\affiliation{%
   \institution{Alexandria University}
  \city{Alexandria}
  \country{Egypt}
}

\author{Sherif Mostafa}
%\authornote{S. Mostafa and M. Youssef are with The American University in Cairo.}
% , Cairo, Egypt.}
\email{sherif\_mostafa@aucegypt.edu}
\orcid{0000-0003-1405-3966}
\affiliation{%
   \institution{The American University in Cairo}
  \city{Cairo}
  \country{Egypt}
}

\author{Moustafa Youssef}
%\authornotemark[2]
\email{moustafa-youssef@aucegypt.edu}
\orcid{0000-0002-2063-4364}
\affiliation{%
   \institution{The American University in Cairo}
  \city{Cairo}
  \country{Egypt}
}

\renewcommand{\shortauthors}{Abdelmotlb and Taman et al.}

\begin{abstract}
Indoor localization systems commonly rely on fingerprinting, which requires extensive survey efforts to obtain location-tagged signal data, limiting their real-world deployability. Recent approaches that attempt to reduce this overhead either suffer from low representation ability, mode collapse issues, or require the effort of collecting data at all target locations.

We present \textit{LocaGen}, a novel spatial augmentation framework that significantly reduces fingerprinting overhead by generating high-quality synthetic data at completely unseen locations. \textit{LocaGen} leverages a conditional diffusion model guided by a novel spatially-aware optimization strategy to synthesize realistic fingerprints at unseen locations using only a subset of seen locations. To further improve our diffusion model performance, \textit{LocaGen} augments seen location data based on domain-specific heuristics and strategically selects the seen and unseen locations using a novel density-based approach that ensures robust coverage. Our extensive evaluation on a real-world WiFi fingerprinting dataset shows that \textit{LocaGen} maintains the same localization accuracy even with 30\% of the locations unseen and achieves up to 28\% improvement in accuracy over state-of-the-art augmentation methods. 
\end{abstract}

\maketitle
% ======================================================================================

% =================== Start Stage 03-introduction ===================
\section{Introduction}
\label{sec:introduction}
With the growing adoption of location-based services, indoor localization systems have recently gained increasing attention like healthcare, emergency response, logistics, and military \cite{mostafa2022SurveyIndoorLocalization, applications_survey}.
Currently, the most prominent indoor localization approach is to rely on signal fingerprinting \cite{mostafa2022SurveyIndoorLocalization}. Fingerprinting depends on collecting \textbf{large amounts} of location-tagged signal data in the environment (e.g., from WiFi access points) to create a ``fingerprint map'' and train a localization model. It then relies on learning location-dependent signal attenuation effects (e.g., due to changing distance from the transmitter and passing through obstructions such as furniture) to allow the localization model to extract location-distinguishing patterns from real-time signal data and infer the location. \textit{Although fingerprinting approaches provide state-of-the-art localization accuracy \cite{mostafa2022SurveyIndoorLocalization}, their practical deployability is severely hindered by the data collection overhead needed to train the localization model \cite{mostafa_unicellular_2023, mostafa2024accurate}.}

Recently, several data augmentation-based approaches have been proposed to reduce the overhead of fingerprint data collection \cite{rizk_monodcell:_2019, shokry_deepcell_2024, tang2024multi, njima_indoor_2021}. However, these approaches either rely on classical machine learning methods \cite{rizk_monodcell:_2019, shokry_deepcell_2024, tang2024multi}, which have low representational ability; 
or employ the commonly used generative adversarial networks (GANs) \cite{njima_indoor_2021} that suffer from mode collapse and training instability problems \cite{mostafa2024UbiquitousLowOverheadFloor}. This causes the available methods to provide synthetic samples from a limited distribution, reducing localization accuracy. Moreover, some of these approaches require collecting data at all the fingerprint locations, necessitating the effort of surveying the whole environment. \textit{This raises the need for an augmentation method capable of providing realistic synthetic samples to ensure high localization accuracy while mitigating the need to survey the entire environment.}

In this paper, we propose \textit{LocaGen}, a novel spatial augmentation framework designed to generate fingerprint data at locations completely unseen during training. By doing so, \textit{LocaGen} can be plugged into any fingerprinting-based localization system to significantly reduce its data collection overhead, since it mitigates the need to survey the entire environment. The main idea behind \textit{LocaGen} is to choose a set of \textit{seen locations} where fingerprint data are collected, use these seen location data to train a conditional deep generative model to synthesize fingerprint data conditioned on the location, and then use it to generate data at \textit{unseen locations}. \textit{LocaGen} then passes the collected seen and generated unseen location data to the localization system to train its location inference model.

\textit{LocaGen} includes modules that address several domain-specific design challenges to ensure realistic data generation at unseen locations for accurate localization with minimal data collection overhead. Specifically, \textit{LocaGen} optimizes the division of fingerprint locations into seen and unseen locations based on a novel density-based approach that ensures that the distribution of the seen location data provides enough coverage for realistic generation of unseen location data. Moreover, \textit{LocaGen} employs heuristic-based domain-specific augmentation methods on seen location data, e.g., through Gaussian noise injection, to increase the amount of training data available to the generative model, avoiding overfitting. Furthermore, in contrast to available augmentation methods that rely on classical machine learning techniques and GANs \cite{rizk_monodcell:_2019, shokry_deepcell_2024, tang2024multi, njima_indoor_2021}, \textit{LocaGen} adapts a state-of-the-art conditional diffusion model \cite{ding_ccdm:_2024} to perform data generation, improving the quality of generated data. Finally, \textit{LocaGen} introduces spatially-aware optimization in training its diffusion model to ensure that the model prioritizes learning the data distribution that allows generating realistic fingerprint data at the target unseen locations.

We implement and evaluate \textit{LocaGen} using a real-world WiFi fingerprinting dataset composed of location-tagged received signal strength (RSS) measurements from \textit{520} WiFi access points collected by \textit{three} users \cite{torres-sospedra_ujiindoorloc_2014}. Our results show that plugging \textit{LocaGen} into a localization system results in the localization accuracy remaining unchanged even with \textbf{30\%} of the locations being unseen. 
Furthermore, it provides 
an improvement of up to \textbf{28\%} compared to adapting the state-of-the-art augmentation methods to generate data in unseen locations \cite{rizk_monodcell:_2019, shokry_deepcell_2024, tang2024multi, njima_indoor_2021}. This highlights the effectiveness of \textit{LocaGen} in significantly reducing data collection overhead in fingerprinting-based systems with minimal reduction in localization accuracy.

The remainder of this paper is structured as follows. Section~\ref{sec_system} provides an overview of \textit{LocaGen}'s framework architecture and discusses the details of the modules that handle different design challenges. Section~\ref{sec_evaluation} presents the evaluation of \textit{LocaGen}. Finally, Section~\ref{sec_conclusion} concludes our paper.

% ======================================================================

\section{The LocaGen Framework}\label{sec_system}
In this section, we present an overview of the \textit{LocaGen} framework, outlining our approach. We then discuss the design challenges faced by \textit{LocaGen} and present the details of the modules addressing them.

% Framework Architecture
\begin{figure}[t]
\centerline{\includegraphics[width=0.78\linewidth]{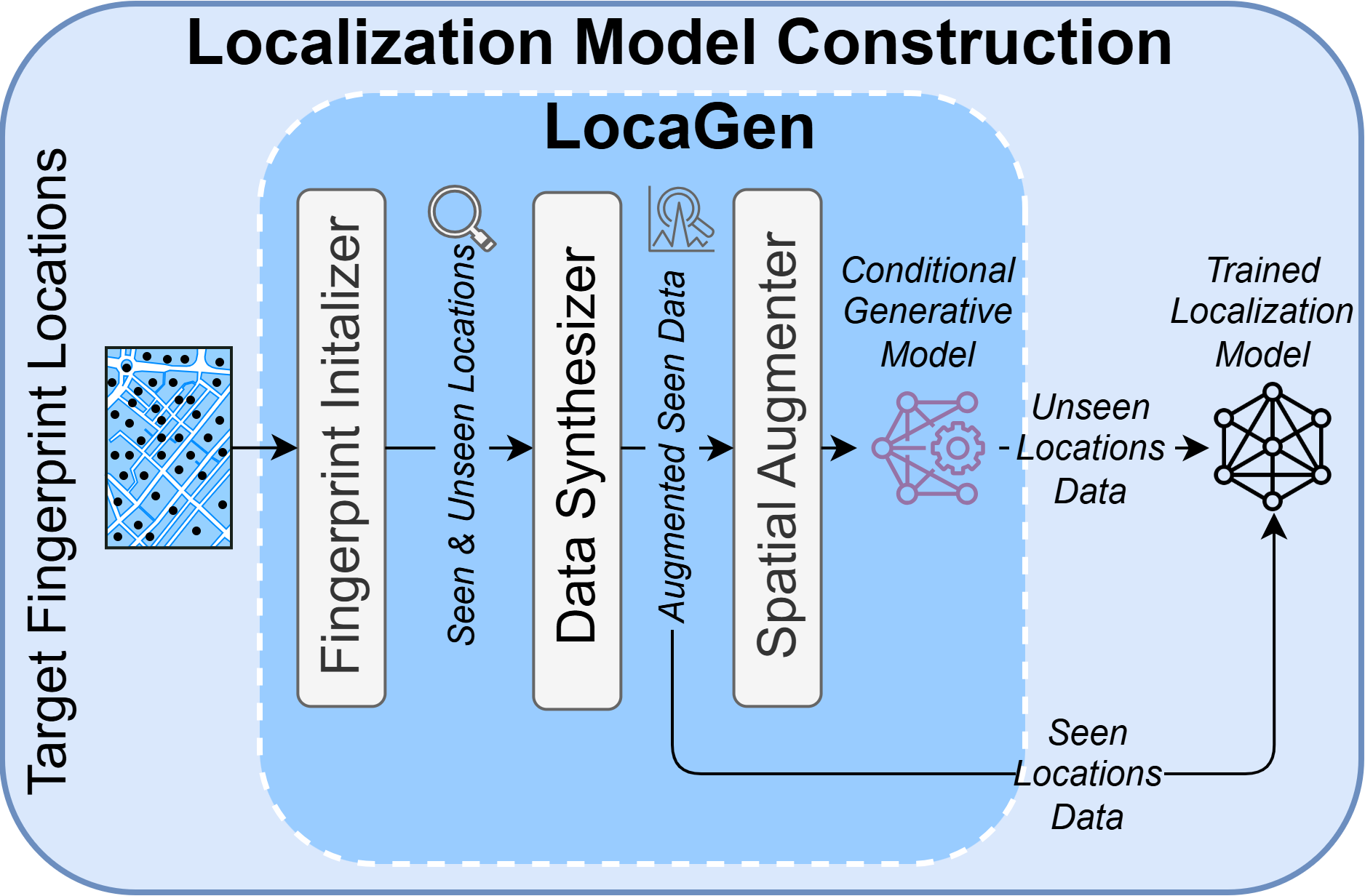}
}
\caption{\textit{LocaGen} Framework Architecture. 
}
\vspace{-13px}
\Description{}{}
\label{fig_architecture}
\end{figure}

\subsection{Framework Overview}
\label{sec_overview}
Figure~\ref{fig_architecture} illustrates the architecture of our \textit{LocaGen} framework, which is designed to be integrated into the data collection process during the construction of the fingerprinting localization model to reduce the collection overhead. \textit{LocaGen} starts with our \textit{Fingerprint Initializer} module dividing the target fingerprint locations into two groups: \textbf{seen locations}, in which fingerprint data are manually collected and constitute the data collection overhead, and \textbf{unseen locations}, whose fingerprint data will be automatically generated by \textit{LocaGen} through spatial augmentation based on modeling the seen location data distribution. 
To ensure the generation of realistic samples in the unseen locations while requiring a minimal number of seen locations, our \textit{Fingerprint Initializer} proposes a density-based approach for its locations' division. After the localization system collects data at the seen locations, our \textit{Data Synthesizer} module 
employs heuristic-based domain-specific data augmentation to create synthetic training data at the seen locations, enhancing \textit{LocaGen}'s generative model ability to learn the fingerprint distribution and generate diverse, realistic samples at the unseen locations. Finally, our \textit{Spatial Augmenter} module trains a conditional generative diffusion model \cite{ding_ccdm:_2024} using data from the seen locations, proposing a novel spatially-aware optimization approach to generate realistic synthetic fingerprints conditioned on the coordinates of the unseen location. Employing a state-of-the-art diffusion model ensures the generation of realistic and diverse data as it mitigates limitations in common generative models, including the mode collapse issues faced by the commonly used GANs \cite{mostafa2024UbiquitousLowOverheadFloor}. 
The augmented seen and generated unseen location data form the full fingerprint map, which is then used to train the localization model.

\subsection{Design Challenges and Framework Details}
\label{sec_challenges_details}
The main challenges \textit{LocaGen} faces include optimizing the selection of seen and unseen locations, the risk of generative model overfitting, and the lack of generative model spatial understanding.

% a. Fingerprint Initializer
\subsubsection{Optimizing Seen Locations Selection}
\label{sec_FingerprintInitializer}
The reliance on data collected at only the seen locations restricts the coverage of the collected signals. This can cause some areas in the environment to be underrepresented and might result in some signal transmitters being completely unseen during training data collection (e.g., when relying on WiFi fingerprinting with limited access point coverage). This makes the selection of seen and unseen locations challenging, as we need to maximize the coverage of collected data to ensure realistic data generation at the unseen locations while minimizing the seen location data collection overhead. 

To address this challenge, our intuition is to prioritize surrounding the unseen locations with a high density of seen locations to capture the needed signal distribution for realistic synthetic data generation. 
To do so, our \textit{Fingerprint Initializer} calculates the average distance from each location to its neighboring locations to quantify the neighbor density for each location in the entire set of target fingerprint locations. 
It then selects those with the highest neighbor density as unseen locations, removes them from the set of target locations, and iteratively repeats the process until the needed number of unseen locations is obtained. The remaining unselected points are designated as seen locations for data collection. 

% b. pre-processor
\subsubsection{The Risk of Generative Model Overfitting}
Since \textit{LocaGen} relies on samples collected from the seen locations only to train its generative model, there is a risk of generative model overfitting. This results in the model generating synthetic data based on a limited data distribution, which reduces the ability of the localization model to learn sufficient location-distinguishing features. 
This is because deep generative models require large amounts of data for training to ensure that they generate samples that reflect the real-world data distributions \cite{mostafa2024accurate}. 

Our \textit{Data Synthesizer} addresses this challenge by augmenting the seen location training data given to our generative model through domain-specific heuristic-based techniques that simulate real-world \textbf{temporal} signal variations. Specifically, it employs \textit{Gaussian noise injection} to generate samples that mimic the effect of dynamic changes in the environment and the noise of the wireless channel on the signal \cite{rizk_effectiveness_2019}. Furthermore, it simulates real-world fluctuations in seen transmitters by dropping transmitters in the fingerprint sample whose signal strength falls below a certain threshold, 
setting their value to zero. 

% c. Loss function modification effect.

 \begin{figure}[t]
\centering
\scalebox{1.2}{
\includegraphics[width=0.8\linewidth]{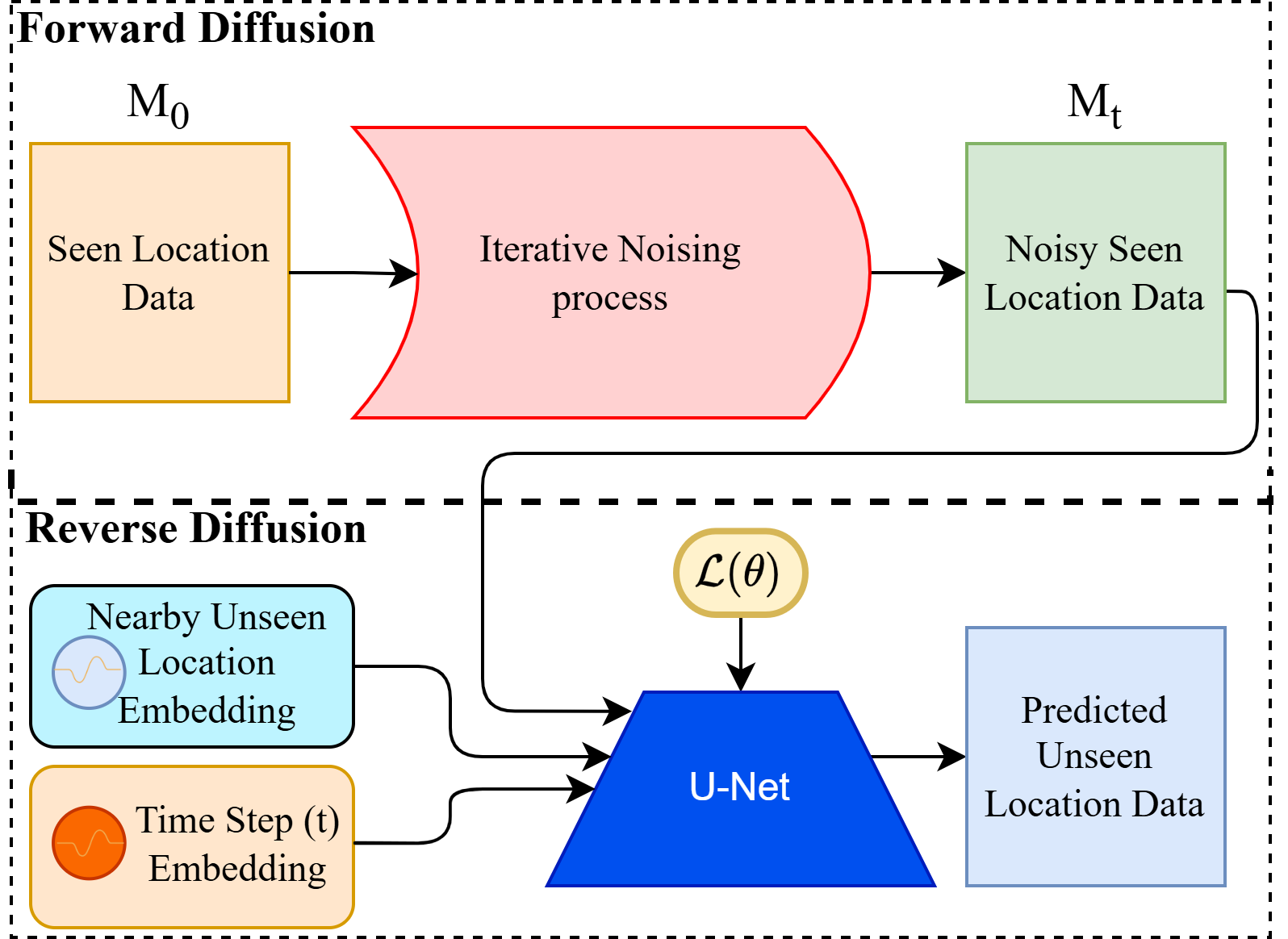}
}
\caption{  
\textit{LocaGen}'s spatial augmentation diffusion-based model. 
The model is optimized with our novel spatially-aware loss function $\mathcal{L}(\theta)$ to ensure synthesizing high-quality samples for the target unseen locations.
}
\vspace{-10px}
\Description{}{}
\label{fig_ccdm} 
\end{figure}

\subsubsection{Our Generative Model}\label{sec_model}
\paragraph{Model Design}
\label{sec_SpatialAugmenter}
    % Model Architecture Diagram
    % background
\textit{LocaGen} introduces a deep generative model design based on the recently proposed continuous conditional diffusion model (CCDM) \cite{ding_ccdm:_2024} to spatially augment the fingerprint data by generating samples in unseen locations \textit{(Figure~\ref{fig_ccdm})}. CCDMs rely on a forward diffusion process to iteratively add noise to the fingerprint signal data $M_0$ at a certain seen location in time steps $t$, generating a noisy version $M_t$. Then, the CCDM trains a U-Net model in a reverse diffusion process to reverse the noise and obtain a prediction of the original data, given a nearby unseen location as a conditional embedding. This allows the U-Net, after training, to generate synthetic data from noise, conditioned on a specific unseen location. To condition the U-Net generation, the location coordinates are embedded and concatenated with the input noisy signal measurements, and the training loss function is weighted based on the distance between the chosen unseen location condition and the ground truth seen location at each training step. 

\paragraph{Integrating Spatial Awareness}The straightforward application of the CCDM samples unseen location condition labels randomly around the ground truth seen location labels \cite{ding_ccdm:_2024}, aiming to cover the entire distribution of possible labels. However, this can cause the model to include impossible unseen location labels in our context of spatial augmentation, where our primary goal is to accurately model the distribution of fingerprint data at specific unseen locations. For example, the CCDM might sample locations completely outside of the building walls, which hinders its learning ability. 

To this end, our \textit{Spatial Augmenter} instead injects spatial awareness into our CCDM by supplying the unseen locations obtained by the \textit{Fingerprint Initializer} directly into the U-Net optimization \textit{(Figure~\ref{fig_ccdm})}. This ensures that the surrounding seen locations are given high weights in the training process to learn their distribution, \textbf{which allows the model to generate high-quality realistic samples at the target unseen locations}, leading to the following spatially-aware loss function for optimizing our U-Net model parameters $\theta$:

    \begin{equation}
    \begin{split}
        \mathcal{L}(\theta) = -\sum_{i=1}^{N_{\text{unseen}}} \sum_{j=1}^{N_\text{seen}} 
        \Bigl[ w(\ell'^{i}, \ell^j)\cdot\| U_{\theta}(t, \ell'^{i}, \mathbf{M}_t^j) - \mathbf{M}_0^j \|^2 \Bigr]
    \end{split}
    \end{equation}
Where $\ell'$ and $\ell$ represent the embeddings of the unseen and seen locations respectively, the distance between which is calculated using $w(\cdot)$ and used as a weight for the loss function $\mathcal{L}(\theta)$. $U_{\theta}$ represents the U-Net model whose parameters are $\theta$, and $\mathbf{M}_t$ and $\mathbf{M}_0$ denote the noisy and non-noisy versions of the signal measurement respectively, obtained at the seen locations $\ell$ at time step $t$.

% ======================================================================
\section{Framework Evaluation}\label{sec_evaluation}

\begin{figure*}[t]
    \centering
    \begin{minipage}{0.24\textwidth}
        \centering
        \scalebox{0.7}{ 
            \input{Diagrams-and-charts/04-Seen-locations-collector/collector}         
        }
        \caption{
            Impact of our Fingerprint Initializer.
        }
        \label{fig:Fingerprint Initializer} 
    \end{minipage}
    \hfill
    \begin{minipage}{0.24\textwidth}
        \centering
        \scalebox{0.7}
        {
            \input{Diagrams-and-charts/06-loss/loss}
        }
        \caption
        {
            Impact of our spatially-aware optimization.
        }
        \label{fig:loss-function}
    \end{minipage}
    \hfill
    \begin{minipage}{0.24\textwidth}
        \centering
        \scalebox{0.65}
        {
            \input{Diagrams-and-charts/07-Overhead/time-overhead}
        }
        \caption
        {
            Effect of the 
            number of seen locations.
        }
        \label{fig:time-overhead}
    \end{minipage}
    \hfill
        \begin{minipage}{0.24\textwidth}
            \centering
            \scalebox{0.53}{  
                \input{Diagrams-and-charts/02-Secondary-Testbed-CDF/second-testbed-cdf.tex}       
            }
            \Description{
                A graph showing the comparison of CDFs for different methods in the second test environment.
            }
            \caption{
                Comparison to state-of-the-art.
            }
            \label{fig:second-testbed-cdf} 
        \end{minipage}
\end{figure*}

\subsection{Dataset and Experimental Setup}

We use a public real-world indoor WiFi fingerprinting dataset \cite{torres-sospedra_ujiindoorloc_2014} to evaluate \textit{LocaGen}, which consists of RSS measurements giving a total of \textit{1600 samples} from \textit{520 WiFi access points} collected by \textit{three users} at \textit{70 locations} along with the corresponding ground truth location coordinates. As our evaluation metric, we implement a localization model based on a fully connected neural network \cite{CellinDeep}, integrating \textit{LocaGen} into its data collection phase, and rely on the mean localization error of this localization model to quantify the impact of \textit{LocaGen}. Unless otherwise specified, we use an unseen-to-seen locations ratio of 50\% in our experiments.

\subsection{Ablation Study}
In this section, we verify the effectiveness of our \textit{Fingerprint Initializer}'s seen location selection approach \textit{(Section~\ref{sec_FingerprintInitializer})} and the impact of our \textit{Spatial Augmenter}'s spatially-aware CCDM training optimization \textit{(Section~\ref{sec_SpatialAugmenter})}. 

\subsubsection{Verifying our Fingerprint Initializer}
\label{eval_FingerprintInitializer}
We compare our \textit{Fingerprint Initializer}'s density-based seen location selection \textit{(Section~\ref{sec_FingerprintInitializer})} with two other methods: the commonly used grid center method, which divides the environment into regions and selects locations at each region's center as seen locations \cite{rizk_monodcell:_2019} and to selecting the seen locations uniformly at random. Figure~\ref{fig:Fingerprint Initializer} shows that \textit{LocaGen}'s density-based selection outperforms grid center selection and random selection \textbf{by over 31\%} in terms of the mean location error.

\subsubsection{Verifying our Spatially-Aware Optimization}
\label{sec:loss-function-effect-experiment}
Figure~\ref{fig:loss-function} shows that our \textit{Spatial Augmenter}'s spatial awareness injection into the loss function of our CCDM \textit{(Section~\ref{sec_SpatialAugmenter})} leads to a decrease in the localization error \textbf{by more than 15\%} compared to directly employing the default CCDM optimization \cite{ding_ccdm:_2024}. 
This is because our spatially-aware optimization enables our CCDM to better focus on optimizing data generation at the target unseen locations. This results in better representation ability for the signal distribution at these locations, leading to high-quality signal data generation.

\subsection{Overall Framework Performance}
In this section, we present a comprehensive evaluation of \textit{LocaGen}'s impact on reducing the data collection overhead of localization systems. 
First, we examine the trade-off between localization accuracy and data collection overhead reduction when employing \textit{LocaGen} with varying unseen-to-seen locations ratio. We then compare \textit{LocaGen} to four state-of-the-art augmentation methods and to the case of using the collected data without spatial augmentation.

\subsubsection{Localization Accuracy-Overhead Trade-Off}
As seen in Figure~\ref{fig:time-overhead}, the data collection overhead decreases linearly with increasing the unseen-to-seen locations ratio, while the mean localization error stays relatively stable at first, \textit{remaining nearly constant until a ratio of 30\%} and showing only a marginal increase up to a ratio of 50\%. Moreover, with a \textit{threefold} reduction in data collection overhead (i.e., from 120 to 40 mins at a 70\% unseen-to-seen locations ratio), the mean localization error increases by \textit{2.24 meters only}. 
This highlights that \textit{LocaGen}'s advantage in reducing the data collection overhead outweighs the increase in localization error. Localization system operators can select the suitable unseen-to-seen locations ratio based on the desired accuracy-overhead trade-off. 
  
\subsubsection{Comparison with State-of-the-Art} 
% a. opening paragraph
We compare \textit{LocaGen} to employing state-of-the-art augmentation techniques to generate data at unseen locations, including a KNN-based spatial interpolator \cite{rizk_monodcell:_2019}, a single-output Gaussian process (SOGP)-based approach \cite{shokry_deepcell_2024}, a multi-output Gaussian process (MOGP)-based method \cite{tang2024multi}, and a selective generative adversarial network (SGAN) approach \cite{njima_indoor_2021}, and to using only the seen locations without augmentation. 
As seen in Figure~\ref{fig:second-testbed-cdf}, \textit{LocaGen} results in superior localization accuracy compared to the state-of-the-art augmentation methods, improving the median location error by 28\%, 27\%, 18\%,  17\%, and 16\%, compared to spatial interpolator \cite{rizk_monodcell:_2019}, no augmentation, MOGP \cite{tang2024multi}, SGAN \cite{njima_indoor_2021}, and SOGP \cite{shokry_deepcell_2024}, respectively. \textit{LocaGen}'s superior performance stems from its use of a state-of-the-art conditional diffusion model \cite{ding_ccdm:_2024} and from its spatially-aware model optimization \textit{(Section~\ref{sec_SpatialAugmenter})}, allowing the generation of high-quality and diverse unseen location data more effectively than other augmentation methods.

\section{Conclusion}
\label{sec_conclusion}
We have presented \textit{LocaGen}, a novel spatial augmentation framework, designed to be plugged into fingerprinting localization systems to significantly reduce data collection overhead. \textit{LocaGen} adapts a state-of-the-art conditional diffusion model to generate realistic fingerprint data at unseen locations based on data from a few seen locations, proposing a novel spatially-aware optimization strategy to improve the quality of the generated fingerprints. Furthermore, \textit{LocaGen} strategically divides the target fingerprint locations into seen and unseen locations to ensure robust coverage and employs domain-specific heuristic-based augmentation to increase the amount of seen location data, reducing overfitting. Our evaluation results on a public dataset demonstrate the effectiveness of \textit{LocaGen} in reducing data collection overhead, maintaining the localization accuracy even with 30\% of the locations unseen, and providing up to 28\% accuracy improvements compared to state-of-the-art augmentation methods. 

\bibliographystyle{ACM-Reference-Format}
\bibliography{bibliography.bib}

\end{document}

%% file: Diagrams-and-charts/04-Seen-locations-collector/collector.tex
% GNUPLOT: LaTeX picture with Postscript
\begingroup
  % Encoding inside the plot.  In the header of your document, this encoding
  % should to defined, e.g., by using
  % \usepackage[cp1252,<other encodings>]{inputenc}
  % \inputencoding{cp1252}%
  \makeatletter
  \providecommand\color[2][]{%
    \GenericError{(gnuplot) \space\space\space\@spaces}{%
      Package color not loaded in conjunction with
      terminal option `colourtext'%
    }{See the gnuplot documentation for explanation.%
    }{Either use 'blacktext' in gnuplot or load the package
      color.sty in LaTeX.}%
    \renewcommand\color[2][]{}%
  }%
  \providecommand\includegraphics[2][]{%
    \GenericError{(gnuplot) \space\space\space\@spaces}{%
      Package graphicx or graphics not loaded%
    }{See the gnuplot documentation for explanation.%
    }{The gnuplot epslatex terminal needs graphicx.sty or graphics.sty.}%
    \renewcommand\includegraphics[2][]{}%
  }%
  \providecommand\rotatebox[2]{#2}%
  \@ifundefined{ifGPcolor}{%
    \newif\ifGPcolor
    \GPcolorfalse
  }{}%
  \@ifundefined{ifGPblacktext}{%
    \newif\ifGPblacktext
    \GPblacktexttrue
  }{}%
  % define a \g@addto@macro without @ in the name:
  \let\gplgaddtomacro\g@addto@macro
  % define empty templates for all commands taking text:
  \gdef\gplbacktext{}%
  \gdef\gplfronttext{}%
  \makeatother
  \ifGPblacktext
    % no textcolor at all
    \def\colorrgb#1{}%
    \def\colorgray#1{}%
  \else
    % gray or color?
    \ifGPcolor
      \def\colorrgb#1{\color[rgb]{#1}}%
      \def\colorgray#1{\color[gray]{#1}}%
      \expandafter\def\csname LTw\endcsname{\color{white}}%
      \expandafter\def\csname LTb\endcsname{\color{black}}%
      \expandafter\def\csname LTa\endcsname{\color{black}}%
      \expandafter\def\csname LT0\endcsname{\color[rgb]{1,0,0}}%
      \expandafter\def\csname LT1\endcsname{\color[rgb]{0,1,0}}%
      \expandafter\def\csname LT2\endcsname{\color[rgb]{0,0,1}}%
      \expandafter\def\csname LT3\endcsname{\color[rgb]{1,0,1}}%
      \expandafter\def\csname LT4\endcsname{\color[rgb]{0,1,1}}%
      \expandafter\def\csname LT5\endcsname{\color[rgb]{1,1,0}}%
      \expandafter\def\csname LT6\endcsname{\color[rgb]{0,0,0}}%
      \expandafter\def\csname LT7\endcsname{\color[rgb]{1,0.3,0}}%
      \expandafter\def\csname LT8\endcsname{\color[rgb]{0.5,0.5,0.5}}%
    \else
      % gray
      \def\colorrgb#1{\color{black}}%
      \def\colorgray#1{\color[gray]{#1}}%
      \expandafter\def\csname LTw\endcsname{\color{white}}%
      \expandafter\def\csname LTb\endcsname{\color{black}}%
      \expandafter\def\csname LTa\endcsname{\color{black}}%
      \expandafter\def\csname LT0\endcsname{\color{black}}%
      \expandafter\def\csname LT1\endcsname{\color{black}}%
      \expandafter\def\csname LT2\endcsname{\color{black}}%
      \expandafter\def\csname LT3\endcsname{\color{black}}%
      \expandafter\def\csname LT4\endcsname{\color{black}}%
      \expandafter\def\csname LT5\endcsname{\color{black}}%
      \expandafter\def\csname LT6\endcsname{\color{black}}%
      \expandafter\def\csname LT7\endcsname{\color{black}}%
      \expandafter\def\csname LT8\endcsname{\color{black}}%
    \fi
  \fi
    \setlength{\unitlength}{0.0500bp}%
    \ifx\gptboxheight\undefined%
      \newlength{\gptboxheight}%
      \newlength{\gptboxwidth}%
      \newsavebox{\gptboxtext}%
    \fi%
    \setlength{\fboxrule}{0.5pt}%
    \setlength{\fboxsep}{1pt}%
    \definecolor{tbcol}{rgb}{1,1,1}%
\begin{picture}(3570.00,2550.00)%
    \gplgaddtomacro\gplbacktext{%
      \csname LTb\endcsname%%
      \put(350,280){\makebox(0,0)[r]{\strut{}$0$}}%
      \csname LTb\endcsname%%
      \put(350,706){\makebox(0,0)[r]{\strut{}$1$}}%
      \csname LTb\endcsname%%
      \put(350,1132){\makebox(0,0)[r]{\strut{}$2$}}%
      \csname LTb\endcsname%%
      \put(350,1557){\makebox(0,0)[r]{\strut{}$3$}}%
      \csname LTb\endcsname%%
      \put(350,1983){\makebox(0,0)[r]{\strut{}$4$}}%
      \csname LTb\endcsname%%
      \put(350,2409){\makebox(0,0)[r]{\strut{}$5$}}%
      \csname LTb\endcsname%%
      \put(915,0){\makebox(0,0){\strut{}Random}}%
      \csname LTb\endcsname%%
      \put(1876,0){\makebox(0,0){\strut{}Grid}}%
      \csname LTb\endcsname%%
      \put(2837,0){\makebox(0,0){\strut{}\textit{LocaGen}}}%
    }%
    \gplgaddtomacro\gplfronttext{%
      \csname LTb\endcsname%%
      \put(133,1344){\rotatebox{-270.00}{\makebox(0,0){\strut{}Mean Location Error (m)}}}%
    }%
    \gplbacktext
    \put(0,0){\includegraphics{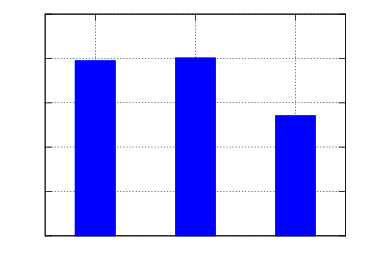}}%
    \gplfronttext
  \end{picture}%
\endgroup

%% file: Diagrams-and-charts/06-loss/loss.tex
% GNUPLOT: LaTeX picture with Postscript
\begingroup
  % Encoding inside the plot.  In the header of your document, this encoding
  % should to defined, e.g., by using
  % \usepackage[cp1252,<other encodings>]{inputenc}
  % \inputencoding{cp1252}%
  \makeatletter
  \providecommand\color[2][]{%
    \GenericError{(gnuplot) \space\space\space\@spaces}{%
      Package color not loaded in conjunction with
      terminal option `colourtext'%
    }{See the gnuplot documentation for explanation.%
    }{Either use 'blacktext' in gnuplot or load the package
      color.sty in LaTeX.}%
    \renewcommand\color[2][]{}%
  }%
  \providecommand\includegraphics[2][]{%
    \GenericError{(gnuplot) \space\space\space\@spaces}{%
      Package graphicx or graphics not loaded%
    }{See the gnuplot documentation for explanation.%
    }{The gnuplot epslatex terminal needs graphicx.sty or graphics.sty.}%
    \renewcommand\includegraphics[2][]{}%
  }%
  \providecommand\rotatebox[2]{#2}%
  \@ifundefined{ifGPcolor}{%
    \newif\ifGPcolor
    \GPcolorfalse
  }{}%
  \@ifundefined{ifGPblacktext}{%
    \newif\ifGPblacktext
    \GPblacktexttrue
  }{}%
  % define a \g@addto@macro without @ in the name:
  \let\gplgaddtomacro\g@addto@macro
  % define empty templates for all commands taking text:
  \gdef\gplbacktext{}%
  \gdef\gplfronttext{}%
  \makeatother
  \ifGPblacktext
    % no textcolor at all
    \def\colorrgb#1{}%
    \def\colorgray#1{}%
  \else
    % gray or color?
    \ifGPcolor
      \def\colorrgb#1{\color[rgb]{#1}}%
      \def\colorgray#1{\color[gray]{#1}}%
      \expandafter\def\csname LTw\endcsname{\color{white}}%
      \expandafter\def\csname LTb\endcsname{\color{black}}%
      \expandafter\def\csname LTa\endcsname{\color{black}}%
      \expandafter\def\csname LT0\endcsname{\color[rgb]{1,0,0}}%
      \expandafter\def\csname LT1\endcsname{\color[rgb]{0,1,0}}%
      \expandafter\def\csname LT2\endcsname{\color[rgb]{0,0,1}}%
      \expandafter\def\csname LT3\endcsname{\color[rgb]{1,0,1}}%
      \expandafter\def\csname LT4\endcsname{\color[rgb]{0,1,1}}%
      \expandafter\def\csname LT5\endcsname{\color[rgb]{1,1,0}}%
      \expandafter\def\csname LT6\endcsname{\color[rgb]{0,0,0}}%
      \expandafter\def\csname LT7\endcsname{\color[rgb]{1,0.3,0}}%
      \expandafter\def\csname LT8\endcsname{\color[rgb]{0.5,0.5,0.5}}%
    \else
      % gray
      \def\colorrgb#1{\color{black}}%
      \def\colorgray#1{\color[gray]{#1}}%
      \expandafter\def\csname LTw\endcsname{\color{white}}%
      \expandafter\def\csname LTb\endcsname{\color{black}}%
      \expandafter\def\csname LTa\endcsname{\color{black}}%
      \expandafter\def\csname LT0\endcsname{\color{black}}%
      \expandafter\def\csname LT1\endcsname{\color{black}}%
      \expandafter\def\csname LT2\endcsname{\color{black}}%
      \expandafter\def\csname LT3\endcsname{\color{black}}%
      \expandafter\def\csname LT4\endcsname{\color{black}}%
      \expandafter\def\csname LT5\endcsname{\color{black}}%
      \expandafter\def\csname LT6\endcsname{\color{black}}%
      \expandafter\def\csname LT7\endcsname{\color{black}}%
      \expandafter\def\csname LT8\endcsname{\color{black}}%
    \fi
  \fi
    \setlength{\unitlength}{0.0500bp}%
    \ifx\gptboxheight\undefined%
      \newlength{\gptboxheight}%
      \newlength{\gptboxwidth}%
      \newsavebox{\gptboxtext}%
    \fi%
    \setlength{\fboxrule}{0.5pt}%
    \setlength{\fboxsep}{1pt}%
    \definecolor{tbcol}{rgb}{1,1,1}%
\begin{picture}(3570.00,2550.00)%
    \gplgaddtomacro\gplbacktext{%
      \csname LTb\endcsname%%
      \put(518,280){\makebox(0,0)[r]{\strut{}$2$}}%
      \csname LTb\endcsname%%
      \put(518,584){\makebox(0,0)[r]{\strut{}$2.2$}}%
      \csname LTb\endcsname%%
      \put(518,888){\makebox(0,0)[r]{\strut{}$2.4$}}%
      \csname LTb\endcsname%%
      \put(518,1192){\makebox(0,0)[r]{\strut{}$2.6$}}%
      \csname LTb\endcsname%%
      \put(518,1497){\makebox(0,0)[r]{\strut{}$2.8$}}%
      \csname LTb\endcsname%%
      \put(518,1801){\makebox(0,0)[r]{\strut{}$3$}}%
      \csname LTb\endcsname%%
      \put(518,2105){\makebox(0,0)[r]{\strut{}$3.2$}}%
      \csname LTb\endcsname%%
      \put(518,2409){\makebox(0,0)[r]{\strut{}$3.4$}}%
      \csname LTb\endcsname%%
      \put(1281,0){\makebox(0,0){\strut{}Default Optimization}}%
      \csname LTb\endcsname%%
      \put(2638,0){\makebox(0,0){\strut{}\textit{LocaGen}}}%
    }%
    \gplgaddtomacro\gplfronttext{%
      \csname LTb\endcsname%%
      \put(133,1344){\rotatebox{-270.00}{\makebox(0,0){\strut{}Mean Location Error (m)}}}%
    }%
    \gplbacktext
    \put(0,0){\includegraphics{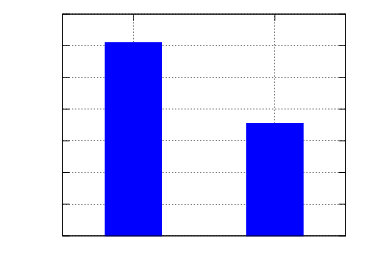}}%
    \gplfronttext
  \end{picture}%
\endgroup

%% file: Diagrams-and-charts/07-Overhead/time-overhead.tex
% GNUPLOT: LaTeX picture with Postscript
\begingroup
  % Encoding inside the plot.  In the header of your document, this encoding
  % should to defined, e.g., by using
  % \usepackage[cp1252,<other encodings>]{inputenc}
  % \inputencoding{cp1252}%
  \makeatletter
  \providecommand\color[2][]{%
    \GenericError{(gnuplot) \space\space\space\@spaces}{%
      Package color not loaded in conjunction with
      terminal option `colourtext'%
    }{See the gnuplot documentation for explanation.%
    }{Either use 'blacktext' in gnuplot or load the package
      color.sty in LaTeX.}%
    \renewcommand\color[2][]{}%
  }%
  \providecommand\includegraphics[2][]{%
    \GenericError{(gnuplot) \space\space\space\@spaces}{%
      Package graphicx or graphics not loaded%
    }{See the gnuplot documentation for explanation.%
    }{The gnuplot epslatex terminal needs graphicx.sty or graphics.sty.}%
    \renewcommand\includegraphics[2][]{}%
  }%
  \providecommand\rotatebox[2]{#2}%
  \@ifundefined{ifGPcolor}{%
    \newif\ifGPcolor
    \GPcolorfalse
  }{}%
  \@ifundefined{ifGPblacktext}{%
    \newif\ifGPblacktext
    \GPblacktexttrue
  }{}%
  % define a \g@addto@macro without @ in the name:
  \let\gplgaddtomacro\g@addto@macro
  % define empty templates for all commands taking text:
  \gdef\gplbacktext{}%
  \gdef\gplfronttext{}%
  \makeatother
  \ifGPblacktext
    % no textcolor at all
    \def\colorrgb#1{}%
    \def\colorgray#1{}%
  \else
    % gray or color?
    \ifGPcolor
      \def\colorrgb#1{\color[rgb]{#1}}%
      \def\colorgray#1{\color[gray]{#1}}%
      \expandafter\def\csname LTw\endcsname{\color{white}}%
      \expandafter\def\csname LTb\endcsname{\color{black}}%
      \expandafter\def\csname LTa\endcsname{\color{black}}%
      \expandafter\def\csname LT0\endcsname{\color[rgb]{1,0,0}}%
      \expandafter\def\csname LT1\endcsname{\color[rgb]{0,1,0}}%
      \expandafter\def\csname LT2\endcsname{\color[rgb]{0,0,1}}%
      \expandafter\def\csname LT3\endcsname{\color[rgb]{1,0,1}}%
      \expandafter\def\csname LT4\endcsname{\color[rgb]{0,1,1}}%
      \expandafter\def\csname LT5\endcsname{\color[rgb]{1,1,0}}%
      \expandafter\def\csname LT6\endcsname{\color[rgb]{0,0,0}}%
      \expandafter\def\csname LT7\endcsname{\color[rgb]{1,0.3,0}}%
      \expandafter\def\csname LT8\endcsname{\color[rgb]{0.5,0.5,0.5}}%
    \else
      % gray
      \def\colorrgb#1{\color{black}}%
      \def\colorgray#1{\color[gray]{#1}}%
      \expandafter\def\csname LTw\endcsname{\color{white}}%
      \expandafter\def\csname LTb\endcsname{\color{black}}%
      \expandafter\def\csname LTa\endcsname{\color{black}}%
      \expandafter\def\csname LT0\endcsname{\color{black}}%
      \expandafter\def\csname LT1\endcsname{\color{black}}%
      \expandafter\def\csname LT2\endcsname{\color{black}}%
      \expandafter\def\csname LT3\endcsname{\color{black}}%
      \expandafter\def\csname LT4\endcsname{\color{black}}%
      \expandafter\def\csname LT5\endcsname{\color{black}}%
      \expandafter\def\csname LT6\endcsname{\color{black}}%
      \expandafter\def\csname LT7\endcsname{\color{black}}%
      \expandafter\def\csname LT8\endcsname{\color{black}}%
    \fi
  \fi
    \setlength{\unitlength}{0.0500bp}%
    \ifx\gptboxheight\undefined%
      \newlength{\gptboxheight}%
      \newlength{\gptboxwidth}%
      \newsavebox{\gptboxtext}%
    \fi%
    \setlength{\fboxrule}{0.5pt}%
    \setlength{\fboxsep}{1pt}%
    \definecolor{tbcol}{rgb}{1,1,1}%
\begin{picture}(3684.00,2776.00)%
    \gplgaddtomacro\gplbacktext{%
      \csname LTb\endcsname%%
      \put(518,588){\makebox(0,0)[r]{\strut{}$0$}}%
      \csname LTb\endcsname%%
      \put(518,832){\makebox(0,0)[r]{\strut{}$20$}}%
      \csname LTb\endcsname%%
      \put(518,1077){\makebox(0,0)[r]{\strut{}$40$}}%
      \csname LTb\endcsname%%
      \put(518,1321){\makebox(0,0)[r]{\strut{}$60$}}%
      \csname LTb\endcsname%%
      \put(518,1566){\makebox(0,0)[r]{\strut{}$80$}}%
      \csname LTb\endcsname%%
      \put(518,1810){\makebox(0,0)[r]{\strut{}$100$}}%
      \csname LTb\endcsname%%
      \put(518,2055){\makebox(0,0)[r]{\strut{}$120$}}%
      \csname LTb\endcsname%%
      \put(518,2299){\makebox(0,0)[r]{\strut{}$140$}}%
      \csname LTb\endcsname%%
      \put(602,448){\makebox(0,0){\strut{}$0$}}%
      \csname LTb\endcsname%%
      \put(873,448){\makebox(0,0){\strut{}$0.1$}}%
      \csname LTb\endcsname%%
      \put(1144,448){\makebox(0,0){\strut{}$0.2$}}%
      \csname LTb\endcsname%%
      \put(1414,448){\makebox(0,0){\strut{}$0.3$}}%
      \csname LTb\endcsname%%
      \put(1685,448){\makebox(0,0){\strut{}$0.4$}}%
      \csname LTb\endcsname%%
      \put(1956,448){\makebox(0,0){\strut{}$0.5$}}%
      \csname LTb\endcsname%%
      \put(2227,448){\makebox(0,0){\strut{}$0.6$}}%
      \csname LTb\endcsname%%
      \put(2497,448){\makebox(0,0){\strut{}$0.7$}}%
      \csname LTb\endcsname%%
      \put(2768,448){\makebox(0,0){\strut{}$0.8$}}%
      \csname LTb\endcsname%%
      \put(3039,448){\makebox(0,0){\strut{}$0.9$}}%
      \put(3123,588){\makebox(0,0)[l]{\strut{}$2$}}%
      \put(3123,832){\makebox(0,0)[l]{\strut{}$3$}}%
      \put(3123,1077){\makebox(0,0)[l]{\strut{}$4$}}%
      \put(3123,1321){\makebox(0,0)[l]{\strut{}$5$}}%
      \put(3123,1566){\makebox(0,0)[l]{\strut{}$6$}}%
      \put(3123,1810){\makebox(0,0)[l]{\strut{}$7$}}%
      \put(3123,2055){\makebox(0,0)[l]{\strut{}$8$}}%
      \put(3123,2299){\makebox(0,0)[l]{\strut{}$9$}}%
    }%
    \gplgaddtomacro\gplfronttext{%
      \csname LTb\endcsname%%
      \put(2377,2629){\makebox(0,0)[r]{\strut{}Time Overhead}}%
      \csname LTb\endcsname%%
      \put(2377,2461){\makebox(0,0)[r]{\strut{}Localization Error}}%
      \csname LTb\endcsname%%
      \put(49,1583){\rotatebox{-270.00}{\makebox(0,0){\strut{}\shortstack{Collection Overhead (mins)}}}}%
      \put(3361,1583){\rotatebox{-270.00}{\makebox(0,0){\strut{}Mean Location Error (m)}}}%
      \put(1820,98){\makebox(0,0){\strut{}\shortstack{Unseen-to-Seen Locations Ratio}}}%
    }%
    \gplbacktext
    \put(0,0){\includegraphics{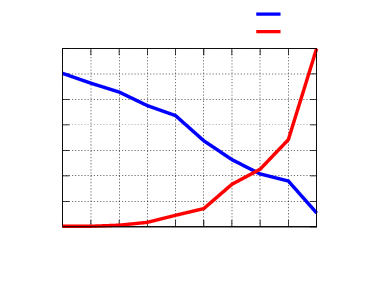}}%
    \gplfronttext
  \end{picture}%
\endgroup

%% file: Diagrams-and-charts/02-Secondary-Testbed-CDF/second-testbed-cdf.tex
% GNUPLOT: LaTeX picture with Postscript
\begingroup
  % Encoding inside the plot.  In the header of your document, this encoding
  % should to defined, e.g., by using
  % \usepackage[cp1252,<other encodings>]{inputenc}
  % \inputencoding{cp1252}%
  \makeatletter
  \providecommand\color[2][]{%
    \GenericError{(gnuplot) \space\space\space\@spaces}{%
      Package color not loaded in conjunction with
      terminal option `colourtext'%
    }{See the gnuplot documentation for explanation.%
    }{Either use 'blacktext' in gnuplot or load the package
      color.sty in LaTeX.}%
    \renewcommand\color[2][]{}%
  }%
  \providecommand\includegraphics[2][]{%
    \GenericError{(gnuplot) \space\space\space\@spaces}{%
      Package graphicx or graphics not loaded%
    }{See the gnuplot documentation for explanation.%
    }{The gnuplot epslatex terminal needs graphicx.sty or graphics.sty.}%
    \renewcommand\includegraphics[2][]{}%
  }%
  \providecommand\rotatebox[2]{#2}%
  \@ifundefined{ifGPcolor}{%
    \newif\ifGPcolor
    \GPcolorfalse
  }{}%
  \@ifundefined{ifGPblacktext}{%
    \newif\ifGPblacktext
    \GPblacktexttrue
  }{}%
  % define a \g@addto@macro without @ in the name:
  \let\gplgaddtomacro\g@addto@macro
  % define empty templates for all commands taking text:
  \gdef\gplbacktext{}%
  \gdef\gplfronttext{}%
  \makeatother
  \ifGPblacktext
    % no textcolor at all
    \def\colorrgb#1{}%
    \def\colorgray#1{}%
  \else
    % gray or color?
    \ifGPcolor
      \def\colorrgb#1{\color[rgb]{#1}}%
      \def\colorgray#1{\color[gray]{#1}}%
      \expandafter\def\csname LTw\endcsname{\color{white}}%
      \expandafter\def\csname LTb\endcsname{\color{black}}%
      \expandafter\def\csname LTa\endcsname{\color{black}}%
      \expandafter\def\csname LT0\endcsname{\color[rgb]{1,0,0}}%
      \expandafter\def\csname LT1\endcsname{\color[rgb]{0,1,0}}%
      \expandafter\def\csname LT2\endcsname{\color[rgb]{0,0,1}}%
      \expandafter\def\csname LT3\endcsname{\color[rgb]{1,0,1}}%
      \expandafter\def\csname LT4\endcsname{\color[rgb]{0,1,1}}%
      \expandafter\def\csname LT5\endcsname{\color[rgb]{1,1,0}}%
      \expandafter\def\csname LT6\endcsname{\color[rgb]{0,0,0}}%
      \expandafter\def\csname LT7\endcsname{\color[rgb]{1,0.3,0}}%
      \expandafter\def\csname LT8\endcsname{\color[rgb]{0.5,0.5,0.5}}%
    \else
      % gray
      \def\colorrgb#1{\color{black}}%
      \def\colorgray#1{\color[gray]{#1}}%
      \expandafter\def\csname LTw\endcsname{\color{white}}%
      \expandafter\def\csname LTb\endcsname{\color{black}}%
      \expandafter\def\csname LTa\endcsname{\color{black}}%
      \expandafter\def\csname LT0\endcsname{\color{black}}%
      \expandafter\def\csname LT1\endcsname{\color{black}}%
      \expandafter\def\csname LT2\endcsname{\color{black}}%
      \expandafter\def\csname LT3\endcsname{\color{black}}%
      \expandafter\def\csname LT4\endcsname{\color{black}}%
      \expandafter\def\csname LT5\endcsname{\color{black}}%
      \expandafter\def\csname LT6\endcsname{\color{black}}%
      \expandafter\def\csname LT7\endcsname{\color{black}}%
      \expandafter\def\csname LT8\endcsname{\color{black}}%
    \fi
  \fi
    \setlength{\unitlength}{0.0500bp}%
    \ifx\gptboxheight\undefined%
      \newlength{\gptboxheight}%
      \newlength{\gptboxwidth}%
      \newsavebox{\gptboxtext}%
    \fi%
    \setlength{\fboxrule}{0.5pt}%
    \setlength{\fboxsep}{1pt}%
    \definecolor{tbcol}{rgb}{1,1,1}%
\begin{picture}(4818.00,3400.00)%
    \gplgaddtomacro\gplbacktext{%
      \csname LTb\endcsname%%
      \put(814,938){\makebox(0,0)[r]{\strut{}$0.1$}}%
      \csname LTb\endcsname%%
      \put(814,1187){\makebox(0,0)[r]{\strut{}$0.2$}}%
      \csname LTb\endcsname%%
      \put(814,1436){\makebox(0,0)[r]{\strut{}$0.3$}}%
      \csname LTb\endcsname%%
      \put(814,1685){\makebox(0,0)[r]{\strut{}$0.4$}}%
      \csname LTb\endcsname%%
      \put(814,1934){\makebox(0,0)[r]{\strut{}$0.5$}}%
      \csname LTb\endcsname%%
      \put(814,2183){\makebox(0,0)[r]{\strut{}$0.6$}}%
      \csname LTb\endcsname%%
      \put(814,2432){\makebox(0,0)[r]{\strut{}$0.7$}}%
      \csname LTb\endcsname%%
      \put(814,2681){\makebox(0,0)[r]{\strut{}$0.8$}}%
      \csname LTb\endcsname%%
      \put(814,2930){\makebox(0,0)[r]{\strut{}$0.9$}}%
      \csname LTb\endcsname%%
      \put(814,3179){\makebox(0,0)[r]{\strut{}$1$}}%
      \csname LTb\endcsname%%
      \put(946,484){\makebox(0,0){\strut{}$0$}}%
      \csname LTb\endcsname%%
      \put(1641,484){\makebox(0,0){\strut{}$5$}}%
      \csname LTb\endcsname%%
      \put(2336,484){\makebox(0,0){\strut{}$10$}}%
      \csname LTb\endcsname%%
      \put(3031,484){\makebox(0,0){\strut{}$15$}}%
      \csname LTb\endcsname%%
      \put(3726,484){\makebox(0,0){\strut{}$20$}}%
      \csname LTb\endcsname%%
      \put(4421,484){\makebox(0,0){\strut{}$25$}}%
    }%
    \gplgaddtomacro\gplfronttext{%
      \csname LTb\endcsname%%
      \put(3722,1537){\makebox(0,0)[r]{\strut{}LocaGen}}%
      \csname LTb\endcsname%%
      \put(3722,1397){\makebox(0,0)[r]{\strut{}SOGP \cite{shokry_deepcell_2024}}}%
      \csname LTb\endcsname%%
      \put(3722,1257){\makebox(0,0)[r]{\strut{}SGAN \cite{njima_indoor_2021}}}%
      \csname LTb\endcsname%%
      \put(3722,1117){\makebox(0,0)[r]{\strut{}MOGP \cite{tang2024multi}}}%
      \csname LTb\endcsname%%
      \put(3722,977){\makebox(0,0)[r]{\strut{}No Augmentation}}%
      \csname LTb\endcsname%%
      \put(3722,837){\makebox(0,0)[r]{\strut{}Spatial Interpolator \cite{rizk_monodcell:_2019}}}%
      \csname LTb\endcsname%%
      \put(341,1941){\rotatebox{-270.00}{\makebox(0,0){\strut{}CDF}}}%
      \put(2683,154){\makebox(0,0){\strut{}Location Error (m)}}%
    }%
    \gplbacktext
    \put(0,0){\includegraphics{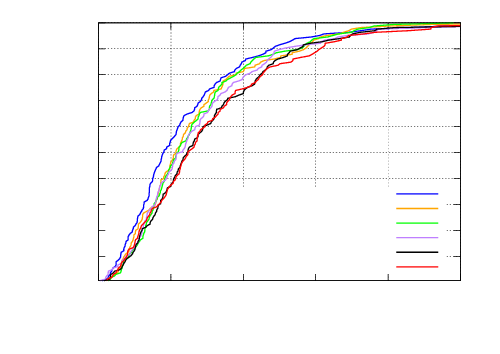}}%
    \gplfronttext
  \end{picture}%
\endgroup